\documentclass{article} 
\usepackage{iclr2024_conference,times}

\usepackage{amsmath,amsthm,amssymb,amsfonts,bbm,bm}

\def\eqref#1{equation~\ref{#1}}

\def\1{\bm{1}}

\DeclareMathAlphabet{\mathsfit}{\encodingdefault}{\sfdefault}{m}{sl}
\SetMathAlphabet{\mathsfit}{bold}{\encodingdefault}{\sfdefault}{bx}{n}

\def\sA{{\mathbb{A}}}

\DeclareMathOperator*{\argmax}{arg\,max}

\usepackage{color,soul}
\usepackage{listings}

\newcommand{\todo}[1]{}
\renewcommand{\todo}[1]{\hl{To-do: #1}}

\def\ie{\textit{i.e.},\ }
\def\eg{\textit{e.g.},\ }

\definecolor{codegray}{rgb}{0.5,0.5,0.5}
\definecolor{backcolour}{rgb}{0.95,0.95,0.95}

\usepackage[hidelinks]{hyperref}
\usepackage{url}

\usepackage{booktabs}
\usepackage{graphicx}
\usepackage{subcaption}
\usepackage{enumitem}

\title{Critique Ability of Large Language Models}

\author{
Liangchen Luo$^\dagger$\thanks{Correspondence: \texttt{\{luolc,leimeng\}@google.com}.}\:\:
Zi Lin$^\ddagger$\,\:
Yinxiao Liu$^\dagger$\,\: 
Lei Shu$^\dagger$\,\:
Yun Zhu$^\dagger$\,\:
Jingbo Shang$^\ddagger$\,\: 
Lei Meng$^\dagger$\footnotemark[1]\\
$^\dagger$Google Research\quad
$^\ddagger$UC San Diego
}

\newcommand{\lzi}[1]{}
\renewcommand{\lzi}[1]{\textcolor{orange}{\textbf{Zi}: #1}}

\newcommand{\ls}[1]{}
\renewcommand{\ls}[1]{\textcolor{magenta}{\textbf{Shu}: #1}}

\newcommand{\yz}[1]{}
\renewcommand{\yz}[1]{\textcolor{blue}{\textbf{Zhu}: #1}}

\newcommand{\lm}[1]{}
\renewcommand{\lm}[1]{\textcolor{pink}{\textbf{leimeng}: #1}}

\renewcommand{\ls}[1]{}
\renewcommand{\yz}[1]{}
\renewcommand{\lm}[1]{}

\iclrfinalcopy 
\begin{document}

\maketitle

\begin{abstract}

Critical thinking is essential for rational decision-making and problem-solving. This skill hinges on the ability to provide precise and reasoned critiques and is a hallmark of human intelligence. In the era of large language models (LLMs), this study explores the ability of LLMs to deliver accurate critiques across various tasks. We are interested in this topic as a capable critic model could not only serve as a reliable evaluator, but also as a source of supervised signals for model tuning. Particularly, if a model can self-critique, it has the potential for autonomous self-improvement. To examine this, we introduce a unified evaluation framework for assessing the critique abilities of LLMs. We develop a benchmark called \textsc{CriticBench}, which comprises $3$K high-quality natural language queries and corresponding model responses; and annotate the correctness of these responses. The benchmark cover tasks such as math problem-solving, code completion, and question answering. We evaluate multiple LLMs on the collected dataset and our analysis reveals several noteworthy insights: (1) Critique is generally challenging for most LLMs, and this capability often emerges only when models are sufficiently large. (2) In particular, self-critique is especially difficult. Even top-performing LLMs struggle to achieve satisfactory performance. (3) Models tend to have lower critique accuracy on problems where they are most uncertain. To this end, we introduce a simple yet effective baseline named \textit{self-check}, which leverages self-critique to improve task performance for various models. We hope this study serves as an initial exploration into understanding the critique abilities of LLMs, and aims to inform future research, including the development of more proficient critic models and the application of critiques across diverse tasks.

\end{abstract}

\section{Introduction}

\begin{quote}
    \textit{``Self-criticism is an art not many are qualified to practice.'' --- Joyce Carol Oates}
\end{quote}

Large language models (LLMs) have demonstrated impressive capacities in a wide range of tasks \citep{Google2023PaLM2,OpenAI2023GPT-4}. Consequently, the evaluation of LLMs has shifted focus from basic sentence coherence to more advanced capabilities, \eg knowledge acquisition and logical reasoning \citep{Hendrycks2021MMLU,BBH2023BigBench}. One capability that is overlooked in current evaluation frameworks is the ability of \textit{critical thinking}, which is an important hallmark of human intelligence that requires logic, reasoning, and knowledge. This ability ensures that LLMs can provide precise and reasoned critiques towards model responses. A model with robust critique ability can identify potential misinformation, errors or context misalignment in model outputs, thereby showing their specific shortcomings that can serve as a feedback for improvement. While recent studies have used LLMs for various forms of critique across diverse applications \citep{Madaan2023SelfRefine,Saunders2022SelfCritique,Shinn2023Reflexion}, they primarily focus on advancing the state of the art for specific tasks instead of providing a comprehensive assessment of critique ability.

\begin{figure}[tb]
  \centering
  \includegraphics[width=\linewidth]{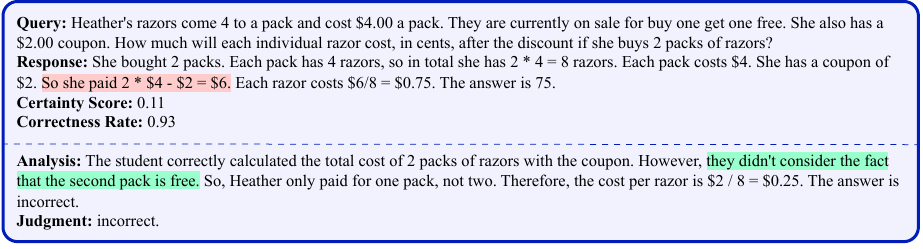}
  \caption{
    An example from \textsc{CriticBench} is presented. The query originates from GSM8K \citep{Cobbe2021GSM8K}, and the response is generated by PaLM-2-L \citep{Google2023PaLM2}. A flaw in the response is highlighted in red. The model shows low confidence in this query, as evidenced by a certainty score of only $0.11$. Below the dashed line, a critique is generated by few-shot prompting PaLM-2-L. It successfully identifies the flaw in the response and makes an accurate judgment. As the policy model and critic model are the same, this example also serves as an instance of \textit{self-critique}.
  }
  \label{fig:intro:data-example}
\end{figure}

To address this gap, we propose a standardized benchmark \textbf{\textsc{CriticBench}} to assess the critique abilities of LLMs in diverse tasks. We define a model's critique ability as ``\textit{the capacity to identify flaws in model responses to queries}''. Figure~\ref{fig:intro:data-example} provides an example of a flaw in the response to a query, and how it is identified by a critique.
The benchmark consists of query-response-judgment triplets. During evaluation, we always prompt a model to perform a chain-of-thought analysis to identify flaws and explain the reason; and then provide a final judgment on the response's correctness. Comparing this judgment to ground-truth labels allows us to \textit{explicitly} evaluate a model's critique accuracy and \textit{implicitly} assess its analytical process toward an accurate judgment.

To construct \textsc{CriticBench} (Section~\ref{sec:data-collection}), we gather natural language queries from multiple scientific benchmarks, covering tasks like math problem-solving~\citep{Cobbe2021GSM8K}, code completion~\citep{Chen2021HumanEval}, and question answering~\citep{Lin2021TruthfulQA}.
We employ PaLM-2 models \citep{Google2023PaLM2} of various sizes to generate responses, which are then annotated for correctness.
To ensure data quality, a complexity-based selection strategy \citep{Fu2023ComplexCoT} is used to identify high-quality responses among the candidates.
Furthermore, to select queries of suitable difficulty, we introduce an auxiliary metric that quantifies a model's certainty regarding a query. Such a metric can help select queries that poses a moderate level of challenge to models.
As a result, we collect $3$K high-quality examples from an initial pool of $780$K candidates to form the benchmark mixture.
This data collection method is both scalable and generalizable, requiring no extra human intervention and suitable for a variety of tasks.

Given \textsc{CriticBench}, we can now analyze the critique abilities of LLMs (Section~\ref{sec:properties-of-critique-ability}). There are specific aspects that particularly interest us. First, critique inherently involves logic, reasoning, and knowledge, making it a complex process even for humans. Therefore, it is not clear how well LLMs can emulate this capability. It is possible that critique ability is yet another emergent ability, \ie ability not present in smaller-scale models that are present in larger-scale models~\citep{Jang2023ReflexionBlog}.
Investigating how critique ability scales with model size could offer insights into model size selection and whether fine-tuning is needed for smaller models  (Section~\ref{sec:scaling-law}).
Additionally, \textit{self-critique}, \ie when a model critiques its own outputs, is a format of critique of particular interest to us, as it is relevant to a model's potential for self-improvement (Section~\ref{sec:self-critique-ability}).
Finally, we are also interested in what types of queries pose more challenges for LLMs to critique (Section~\ref{sec:correlation-to-certainty}).

To investigate these aspects, we evaluate various widely-used LLMs on \textsc{CriticBench} and reveal several intriguing findings:
(1) Critique tasks pose a considerable challenge for LLMs. Only large-scale models exhibit performance with a notable difference from a random guess baseline, indicating that the capacity for critique serves as an emergent indicator of a capable LLM.
(2) Self-critique, \ie a model critiquing its own output, is particularly difficult. Even the strongest LLMs struggle to achieve satisfactory performance.
(3) A challenging query is not only difficult for LLMs to directly answer correctly, but also poses a challenge in assessing an answer's correctness to that query.

To this end, we also propose a simple yet effective baseline called \textit{self-check} (Section~\ref{sec:application-of-critique}). The basic idea is to prompt the model to confirm the accuracy of their generated answers by self-critique before presenting them. The method consistently enhances the baseline performance~\citep{Wang2023SelfConsistency} on math word problems across multiple models, achieving an average of 9.55\% error reduction rate, which demonstrates the potential utility of critiques from LLMs.

Our contributions are three-fold:
\begin{itemize}[nosep,leftmargin=*]
\item \textbf{New Benchmark}\quad
\textsc{CriticBench} is the first benchmark that comprehensively assesses the critique abilities of LLMs across diverse tasks and scenarios, which fills a gap in the current LLM evaluation framework by introducing this important ability.

\item \textbf{New Findings}\quad
Our findings on \textsc{CriticBench} underscore the nuances and depth of LLM's critique abilities (Section~\ref{sec:properties-of-critique-ability}). These revelations enhance our understanding of the inherent complexities in LLMs and emphasize the need for advanced training and evaluation techniques.

\item \textbf{New Capacity}\quad
The proposed \textit{self-check} method (Section~\ref{sec:application-of-critique}) not only advances the performance on math word problems over the baseline, but also indicates the new capacity of critique ability with LLMs, which is a fruitful avenue for LLM's self-improvement strategies.
\end{itemize}

\section{Definition of Critique Ability}
\label{sec:definition-of-critique-ability}

The concept of \textit{critique} has diverse interpretations and is often applied informally in everyday contexts. Recent research employs large language models to offer critiques across multiple applications~\citep{Madaan2023SelfRefine,Paul2023Refiner,Saunders2022SelfCritique,Shinn2023Reflexion}, resulting in varying formats and requirements for their ``critiques''. These studies primarily aim to enhance performance in specific tasks, neglecting to clarify the meaning of the term critique. In this paper, we consider the definition of a language model's critique ability as
\begin{center}
    \textit{the capacity to identify flaws in model responses to queries.}
\end{center}
These flaws can differ depending on the task, ranging from incorrect reasoning or calculation in mathematical problems to syntax errors in code completion.

When a model self-assesses its own outputs, we term this as \textit{self-critique}, a notion that particularly intrigues us. If models can engage in self-critique and reflection, they can potentially do self-improvement, requiring minimal human intervention.
On the risky side, this autonomy also raises concerns about reduced human oversight \citep{Bowman2022OversightLLM}. Yet we posit that self-critique may still remain a challenging capability for large language models, as a flaw-aware model would logically not produce faulty output in the first place \citep{Saunders2022SelfCritique}.



\section{Construction of \textsc{CriticBench}}
\label{sec:data-collection}
\ls{after reading data collection, I have a wired feeling. Your critique data collection from PaLM2 and selected arbitrarily to form critic bench. Then you evaluate critic ability on PaLM 2 models again. Does your critic ability analysis try to reflect your arbitrary critic-data selection? The llama scores are close to random but PaLMs show emergent scaling-law, is this the reflection of your arbitrary critic bench from PaLM 2. Furthermore, as you use truthful qa to demonstrate that model belief (false) ~= model answer (false) ~= model critic (false), do we have other proof on belief (true) ~= model answer (true / false) ~= model critic (true / false) ? }
As discussed in Section~\ref{sec:definition-of-critique-ability}, prior research employs large language models to offer critiques, yet requires particular process and formats to meet their task-specific objectives. Currently, there is no standard or generalizable way to assess the critique abilities of language models across diverse tasks. This section proposes \textsc{CriticBench}, a unified, standardized evaluation framework to tackle this issue. The framework aims to fulfill three criteria:

\begin{itemize}[nosep,leftmargin=*]
\item \textbf{Scalability}\quad
Given the broad range of tasks already established within the community, and the anticipation of more to emerge, a scalable data collection method is essential. The method should minimize human annotation efforts and ideally be fully autonomous.

\item \textbf{Generalizability}\quad
The framework should be task-agnostic, capable of generalizing across various tasks and domains.

\item \textbf{Quality}\quad
We believe quality matters more than quantity. When volume of data is substantial, we prioritize selecting those that most effectively differentiate between stronger and weaker models.
\end{itemize}

The following subsections illustrate the detailed construction process. Specifically, Section~\ref{sec:data-generation} presents the initial data generation on three different tasks, where we get the collection of query-response-judgment triplets as shown in Figure~\ref{fig:intro:data-example}. Section~\ref{sec:data-selection} then shows how to select data based on the initial collection to guarantee the quality of responses and queries.

\subsection{Data Generation}
\label{sec:data-generation}
For the tasks of interest, we begin by employing existing scientific datasets from relevant domains. These datasets are expected to include queries that large language models, which here we refer to as \textit{generators}, aim to respond. 

To ensure scalability, it is essential to have an automated approach for assessing the correctness of a model's responses. Classification tasks naturally meet this criterion, as model outputs can be automatically compared to ground-truth labels. Similarly, tasks that involve auto-verifiable answers also comply; for instance, in code completion tasks with unit tests available, the validity of the generated code can be confirmed by passing all tests. For free-form generation tasks such as summarization and translation, assessing the quality of a response remains non-trivial. However, recent advances in LLM-based automated evaluation for generation tasks mitigate this issue to some extent, enabling the assessment without human intervention \citep{Liu2023GEval}. 

While not exhaustive, these already cover a significant range of tasks and domains. We acknowledge the limitations in some auto-assessment approaches, especially for generation tasks. Improving the reliability of these automated evaluation methods, however, is beyond the scope of this paper.

We employ five different sizes of PaLM-2 models \citep{Google2023PaLM2} as our generators. These models are pretrained solely for next-token prediction and do not undergo supervised fine-tuning or reinforcement learning from human feedback. For coding-related tasks, apart from the standard PaLM-2 models, we also employ the specialized PaLM-2-S* variant. The latter is obtained through continual training of PaLM-2-S on a data mixture enriched with code-heavy corpus.

\textbf{Query Collection}\quad
We extract queries from three datasets: GSM8K \citep{Cobbe2021GSM8K}, HumanEval \citep{Chen2021HumanEval}, and TruthfulQA \citep{Lin2021TruthfulQA}, covering the tasks of math-problem solving, code completion and question answering. For datasets with distinct training and test splits, we use the test data; for datasets intended only for evaluation, all examples are used. Detailed considerations and rationale behind the selection of these datasets are provided in Appendix~\ref{appendix:sec:data-source}.

\textbf{Response Generation}\quad 
We sample $k$ responses for each query, with $k=64$ for GSM8K and TruthfulQA, and $k=100$ for HumanEval. In the case of TruthfulQA, we employ its multiple-choice variation to facilitate autonomous answer annotation.
After filtering out invalid outputs such as empty ones, we collect a total of $780$K responses as an initial pool of candidates. 

\textbf{Annotation for Correctness}\quad 
For GSM8K, we assess answer correctness by comparing its numeric equality to the ground truth, as described by \citet{Lewkowycz2022Minerva}. For HumanEval, correctness is determined by the passage of provided unit tests. For TruthfulQA, we utilize its classification format, judging correctness based on a match with the ground-truth label. 

More details on hyper-parameter settings and prompt templates are available in Appendix~\ref{appendix:sec:data-generation-settings}. 

\subsection{Data Selection}
\label{sec:data-selection}

Many existing evaluation benchmarks for large language models suffer from insufficient differentiability, \ie both stronger and weaker models yield similar performance \citep{Fu2023ReviewOfITBlog}. This issue likely arises from the presence of either overly simple or exceedingly difficult examples in the benchmarks. Such examples are less valuable for evaluation and can undermine the utility of the benchmarks when average scores are calculated, leading to indistinguishable outcomes. 
To address the issue, we introduce various filtering strategies aimed at selecting high-quality and differentiable examples.

\subsubsection{High-Quality Response Selection}

Initially, we can narrow the example set from $780$K to $15$K by sampling one correct and one incorrect response for each query and generator. While random uniform sampling is the the most straightforward strategy, it risks including examples with obvious errors, which offer little evaluative value.  To mitigate this, for the incorrect responses we focus on sampling \textit{convincing wrong-answers} \citep{Lightman2023PRM800K} that are more likely to fool the models.
In cases suitable for majority voting, we identify the most frequent incorrect answer for each query, and then sample from responses containing this answer. For coding tasks where majority voting is not applicable, we sample from responses that pass the most unit tests, indicating that it is mostly correct but fails in certain corner cases. 

To further enhance data quality, we employ the \textit{complexity-based} sample selection strategy \citep{Fu2023ComplexCoT} for tasks that require chain-of-thought reasoning. Specifically, we opt for responses that involve more reasoning steps, as this is positively correlated with higher accuracy \citep{Fu2023ComplexCoT}. This approach is beneficial for sampling both types of responses. For correct ones, it minimizes the likelihood of false positives; for incorrect ones, it yields more convincing responses that pose greater challenges in flaw detection for weaker models.

Lastly, as many tasks are challenging and may require emergent abilities \citep{Wei2022EmergentAbility} to perform well, smaller models generally underperform and produce lower-quality responses compared to larger ones. We include data from these smaller models only for analyzing self-critique abilities; they are excluded from the final evaluation benchmarks.

\subsubsection{Certainty-Based Query Selection}
\label{sec:certainty-based-selection}

Thus far, our focus has been on choosing responses with higher quality and likelihood of accuracy. We now shift our focus to the quality of queries. Not all queries are equally valuable: trivial queries that models easily answer correctly are generally less valuable, whereas queries consistently answered incorrectly may either be too complex for LLMs or suffer from wrong ``golden'' labels.


To minimize the presence of such queries in our benchmark, we introduce two metrics to quantify the levels of certainty when models answer a query: the \textbf{\textit{certainty score}} and \textbf{\textit{correctness rate}}. We will use these metrics to help us select queries of moderate difficulty.

The metrics draw inspiration from the majority voting mechanism in the \textit{self-consistency} approach \citep{Wang2023SelfConsistency}, which functions by generating multiple candidate outputs for a query, and then aggregating them using a majority voting procedure to select the most commonly occurring answer. Observing that different majority votes, even those resulting in the same outcome, can indicate vastly different levels of certainty. To illustrate, consider a voting situation with $100$ candidates where: (i) all candidates are $x$; and (ii) $51$ candidates are $x$ and $49$ are $y$. Although both situations result in a majority vote for $x$, the level of certainty varies significantly: the former situation denotes a high degree of confidence, whereas the latter reflects a considerable level of uncertainty.

Motivated by the observations above, we propose the following method to measure levels of uncertainty in language model responses.
Suppose we prompt a language model $\mathrm{LM}: P(a|q)$ with a query $q$ and sample a bag of $k$ answers $\sA = \{a_i\}_{i=1}^k$, where $a_i \sim P(a|q)$. We denote the most and the second most frequent answers among these $k$ responses as $a^{(1)}$ and $a^{(2)}$, respectively. Uncertainty is measured by the frequency ratio of $a^{(2)}$ to $a^{(1)}$, where a larger ratio indicates a higher level of uncertainty. We term this ratio as \textit{uncertainty rate}. An uncertainty rate of $1$ --- where the two most frequent answers appear with equal frequency --- indicates extremely high model uncertainty. Conversely, an uncertainty rate of $0$, implying that $a^{(2)} = 0$, suggests that all responses are consistent, indicating the model’s strong confidence in its answer.

Formally, we use
$
    f_\sA(a) = \sum_{a_i \in \sA} \1_\mathrm{condition} \left( a_i = a \right)
$
to denote the frequency of an answer $a$ among a bag of responses $\sA$ and
$
    \mathrm{mode}(\sA) = \argmax_a f_\sA(a)
$
to denote the \textit{mode}, \ie the most frequently occurring item \yz{why use this notation?}, of $\sA$.
The uncertainty rate over model responses $\sA$ to the query $q$ is then defined as
$
    \mathrm{UR}_\mathrm{LM}(q;k) = \frac{f_\sA(\mathrm{mode}(\sA \backslash \sA^{(1)}))}{f_\sA(\mathrm{mode}(\sA))}
$,
where
$\sA^{(1)} = \left\{ a \mid a = \mathrm{mode}(\sA), a \in \sA \right\}$
represents the most frequent responses in $\sA$.
For the sake of conciseness and readability in our subsequent discussion, we also define a metric by the negative logarithm of the uncertainty rate to measure model certainty, represented as
$
    \mathrm{Certainty}_\mathrm{LM}(q;k) = -\log \left( \mathrm{UR}_\mathrm{LM}(q;k) \right)
$,
where a larger value indicates a higher level of certainty.
We term it as the \textbf{\textit{certainty score}}.

\ls{how to calculate the uncertainty of generation tasks, sample multiple answers (top-2?) and using probability scores or use the original occurrence?}
In cases where the expected correct answer to a query is available, such as during model evaluation on a test dataset, the definitions above can be slightly adapted to introduce a new metric called \textbf{\textit{correctness rate}}. This metric is defined as the frequency ratio of the correct answer to the most common wrong answer:
$
    \mathrm{CR}_\mathrm{LM}(q;k) = \frac{f_\sA(a^{(\text{e})})}{f_\sA(\mathrm{mode}(\sA_\text{wrong}))}
$,
\yz{What is the motivation for this metrics. What is the diff with accuracy?}
where $a^{(\text{e})}$ denotes the expected answer and
$\sA_\text{wrong} = \left\{ a \mid a \neq a^{(\text{e})}, a \in \sA \right\}$
denotes the incorrect responses.
Using self-consistency, the model votes a correct answer when the correctness rate exceeds $1$, and conversely, it produces an incorrect answer when the rate falls below $1$. In addition, as the rate approaches $1$, the model exhibits increasing levels of uncertainty regarding the answer, no matter if it is correct or not.
This metric naturally reflects the difficulty of a query to the model.

\yz{Will certainty change a lot with different sampling temperature?}

\begin{figure}[htb]
    \centering
    \begin{subfigure}{0.33\textwidth}
        \centering
        \includegraphics[width=\linewidth]{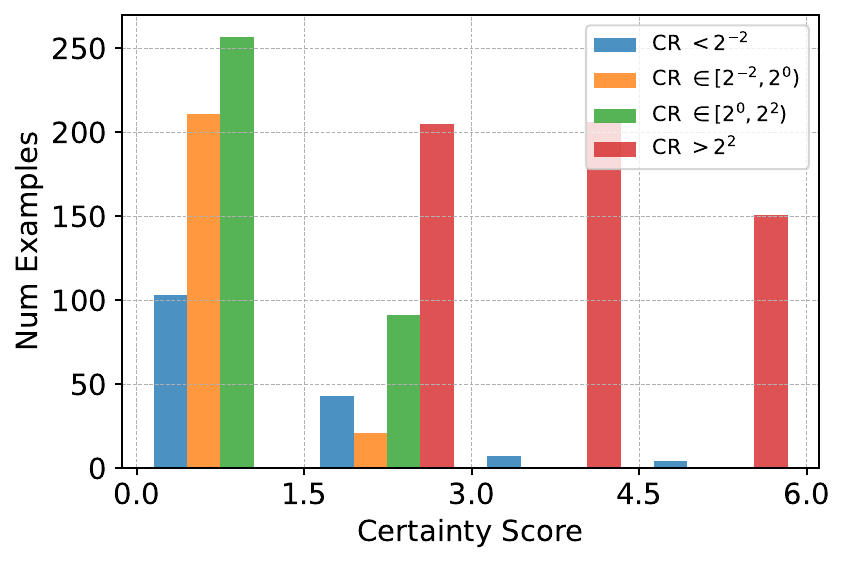}
        \caption{Relation of certainty to $\mathrm{CR}$.}
        \label{fig:certainty:case-study:gsm8k-test-ulm24b:scatter-certainty-to-cr}
    \end{subfigure}
    \begin{subfigure}{0.365\textwidth}
        \centering
        \includegraphics[width=\linewidth]{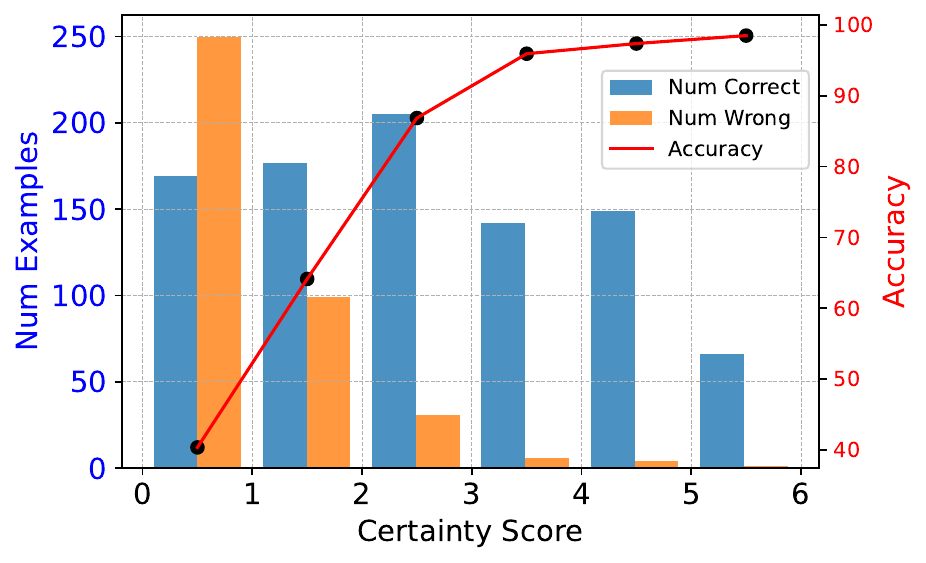}
        \caption{Relation of certainty to accuracy.}
        \label{fig:certainty:case-study:gsm8k-test-ulm24b:certainty-to-accuracy}
    \end{subfigure}
    \caption{Certainty of PaLM-2-S on GSM8K: Relation to correctness rate ($\mathrm{CR}$) and accuracy; based on the $8$-shot chain-of-thought prompt from \citet{Wei2022ChainOfThought} and a $64$-path self-consistency.}
    \label{fig:certainty:case-study:gsm8k-test-ulm24b}
\end{figure}

We present a simple case study to intuitively demonstrate the properties of our proposed metrics. We evaluate PaLM-2-S \citep{Google2023PaLM2} on GSM8K \citep{Cobbe2021GSM8K} using 
a $64$-path self-consistency. The relationship between model certainty, correctness rate ($\mathrm{CR}$), and model accuracy is depicted in Figure~\ref{fig:certainty:case-study:gsm8k-test-ulm24b}.

Figure~\ref{fig:certainty:case-study:gsm8k-test-ulm24b:scatter-certainty-to-cr} displays the correlation between model certainty and correctness rate ($\mathrm{CR}$). Test examples with lower $\mathrm{CR}$ present greater challenges to models. As evidenced in the figure, lower certainty correlates with more low-$\mathrm{CR}$ examples, leading to more incorrect predictions. As certainty increases, the instances of low $\mathrm{CR}$ diminish, resulting in higher accuracy.
Figure~\ref{fig:certainty:case-study:gsm8k-test-ulm24b:certainty-to-accuracy} illustrates the correlation between model certainty and accuracy in a more straightforward way. As the certainty level rises, the proportion of incorrect predictions markedly decreases, signifying increased accuracy.

We now adopt a \textit{certainty-based} sample selection strategy. We calculate the correctness rate for each query, selecting those with a $\mathrm{CR}$ close to $1$. This suggests that models exhibit considerable hesitation and uncertainty when responding to these queries, indicating a moderate level of difficulty that is neither excessively simple ($\mathrm{CR} \rightarrow +\infty$) nor overly challenging ($\mathrm{CR} \rightarrow 0$). For coding tasks, where certainty metrics cannot be computed, we use the ratio of correct to incorrect answers as a surrogate for $\mathrm{CR}$. Moreover, due to the limited size of HumanEval, we only exclude the simpler queries with a $\mathrm{CR} > 1$, and retain the challenging examples. We will analyze the correlation between critique ability and model certainty for queries in Section~\ref{sec:correlation-to-certainty}.

Detailed implementation of each stage in data selection can be found in Appendix~\ref{appendix:sec:data-selection-details}.


\textbf{Final Data Formulation}\quad
To this end, we could further narrow the benchmark dataset to $3$K high-quality, differentiable examples, with $1$K for each original dataset\ls{what is the good/bad distribution? half-half?}. The resulting subsets are named as Critic-GSM8K, Critic-HumanEval, and Critic-TruthfulQA, and their mixture is referred to as \textbf{\textsc{CriticBench}}. We provide the data statistics and examples in Appendix~\ref{appendix:sec:data-statistics-and-examples}. As our data collection method is scalable and generalizable across tasks, we view the construction of \textsc{CriticBench} as a continuous effort. This paper serves as an initial step, presenting three representative datasets. We hope to extend the mixture to cover more tasks and scenarios in future work.

\section{Properties of Critique Ability}
\label{sec:properties-of-critique-ability}

In this section, we conduct our analysis of the critique ability of large language models on \textsc{CriticBench}. We focus primarily on the following three aspects: (1) how critique ability scales with model size (Section~\ref{sec:scaling-law}); (2) models' self-critique ability (Section~\ref{sec:self-critique-ability}); and (3) the correlation between critique ability and models' certainty in response to a query (Section~\ref{sec:correlation-to-certainty}).

For each query-response pair in the dataset, we employ few-shot prompting to instruct models to first conduct a chain-of-thought analysis to identify any flaws in the response and explain the reason; and subsequently issue a judgment on the response's correctness. In evaluation, we focus solely on the accuracy of this final judgment, disregarding the correctness of the intermediate analysis. As empirical evidence has shown a strong correlation between the accuracy of intermediate chain-of-thought and the final answer \citep{Wei2022ChainOfThought,Lewkowycz2022Minerva,Fu2023CoTHub}, we can use the final judgment accuracy as a proxy for the model's critique analysis capability.
Details about the evaluation settings can be found in Appendix~\ref{appendix:sec:eval-settings}.

\subsection{Scaling Law}
\label{sec:scaling-law}

\begin{figure}[htb]
    \centering
    \begin{subfigure}{0.32\textwidth}
        \centering
        \includegraphics[width=\linewidth]{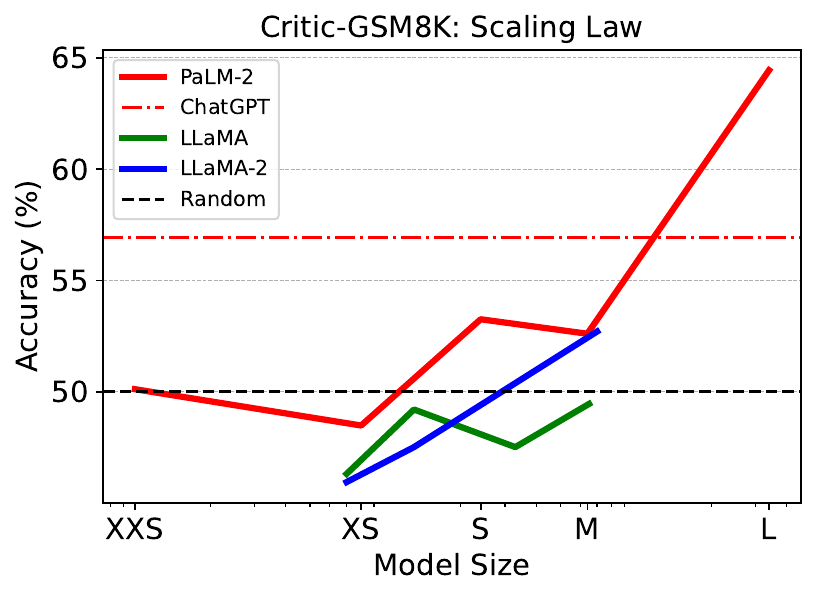}
    \end{subfigure}
    \hfill
    \begin{subfigure}{0.32\textwidth}
        \centering
        \includegraphics[width=\linewidth]{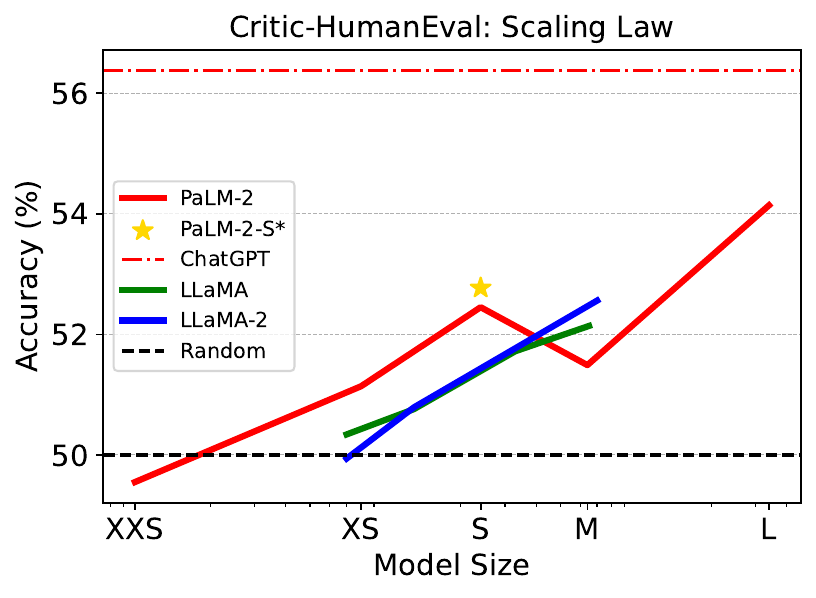}
    \end{subfigure}
    \hfill
    \begin{subfigure}{0.32\textwidth}
        \centering
        \includegraphics[width=\linewidth]{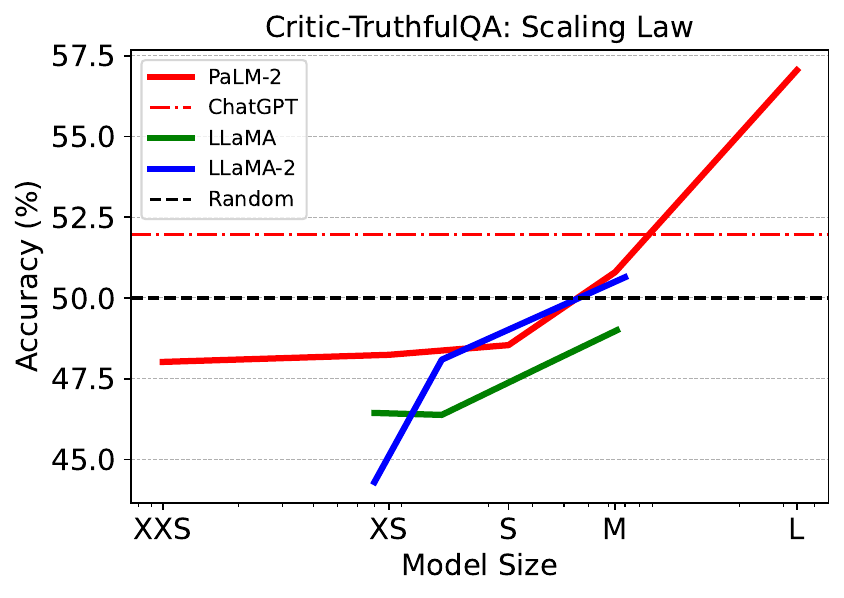}
    \end{subfigure}
    \caption{Scaling law of critique ability: Following \citet{Google2023PaLM2}, we use T-shirt size notations to denote model sizes. All medium-sized or smaller models exhibit poor performance on all tasks, akin to random guessing. Critic-HumanEval poses a great challenge for all models.}
    \label{fig:scaling-law:accuracy}
\end{figure}

\citet{Jang2023ReflexionBlog} posits that critique ability may be an emergent ability \citep{Wei2022EmergentAbility} that only emerges at certain scales of model size. We emphasize that it is better to seek an answer to this hypothesis before directing our efforts toward the applications of critiques. For a critic model to successfully improve the performance of specific tasks, it must possess at least moderate effectiveness. It is possible that the critique ability of smaller models is as futile as a random guess, rendering them incapable for downstream applications. A study of the \textit{scaling law} of critique ability could provide us insights into the appropriate model size selection and whether fine-tuning should be considered for smaller models.

We evaluate multiple widely-used LLM families available in various sizes on \textsc{CriticBench}, including PaLM-2 \citep{Google2023PaLM2}, LLaMA \citep{Touvron2023LLaMA}, LLaMA-2 \citep{Touvron2023LLaMA2}, and ChatGPT \citep{OpenAI2023GPT-4}. Figure~\ref{fig:scaling-law:accuracy} illustrates the scaling behavior of their critique abilities.
The results for ChatGPT are not directly comparable to those of other models because its size is not disclosed and it undergoes instruction-tuning, whereas the others are all pretrained models. We include it here solely for reference purpose.
On Critic-GSM8K and Critic-TruthfulQA, all models of medium size or smaller exhibit poor performance, akin to random guessing. Only PaLM\nobreakdash-2\nobreakdash-L demonstrates non-trivial better results. On Critic-HumanEval, all models perform poorly; even the strongest pretrained model, PaLM-2-L, only achieves an accuracy score of merely $54.14\%$, which is just marginally better than a random guess. This is somewhat anticipated, as evaluating the correctness of a code snippet without execution is often challenging even for expert software engineers. It is likely to gain a notable improvement when augmented by a code interpreter tool. Thus, the benchmark also serves as an ideal testbed to assess LLMs' tool-use capability.

The observed scaling law supports the emergent ability hypothesis by \citet{Jang2023ReflexionBlog}. It suggests that the ability of critique is yet another key indicator of a strong large language model.

\subsection{Self-Critique Ability}
\label{sec:self-critique-ability}

\begin{figure}[htb]
    \centering
    \begin{subfigure}{0.32\textwidth}
        \centering
        \includegraphics[width=\linewidth]{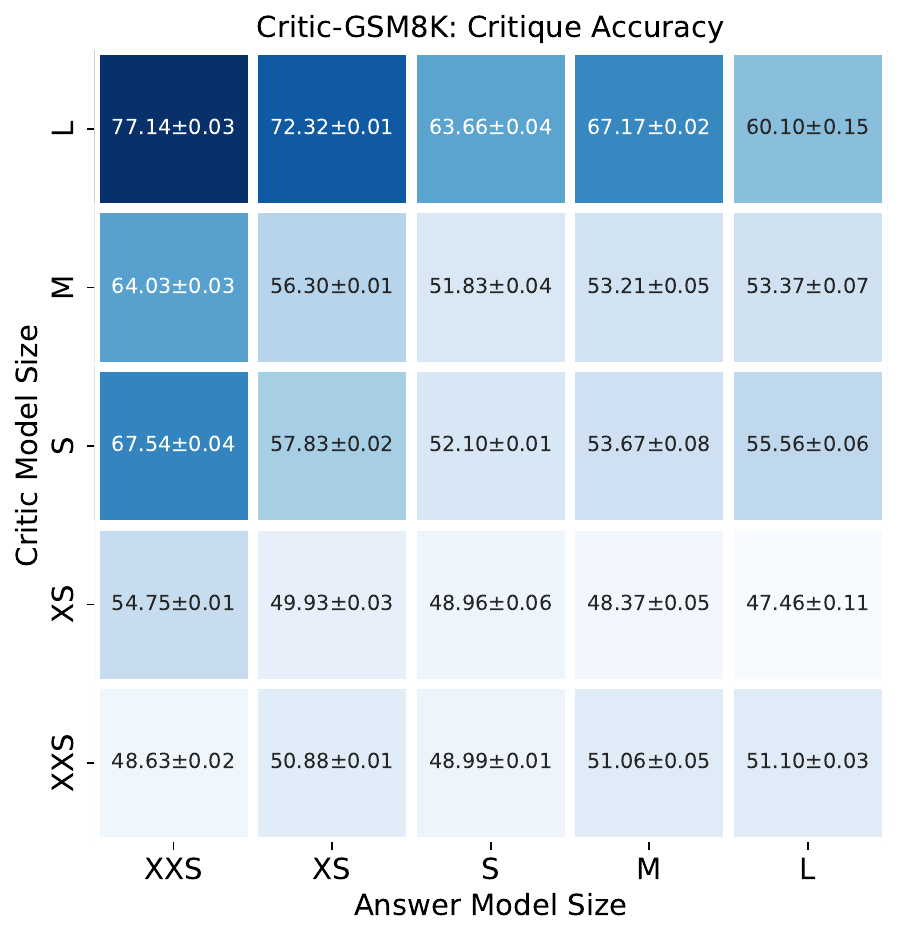}
    \end{subfigure}
    \hfill
    \begin{subfigure}{0.32\textwidth}
        \centering
        \includegraphics[width=\linewidth]{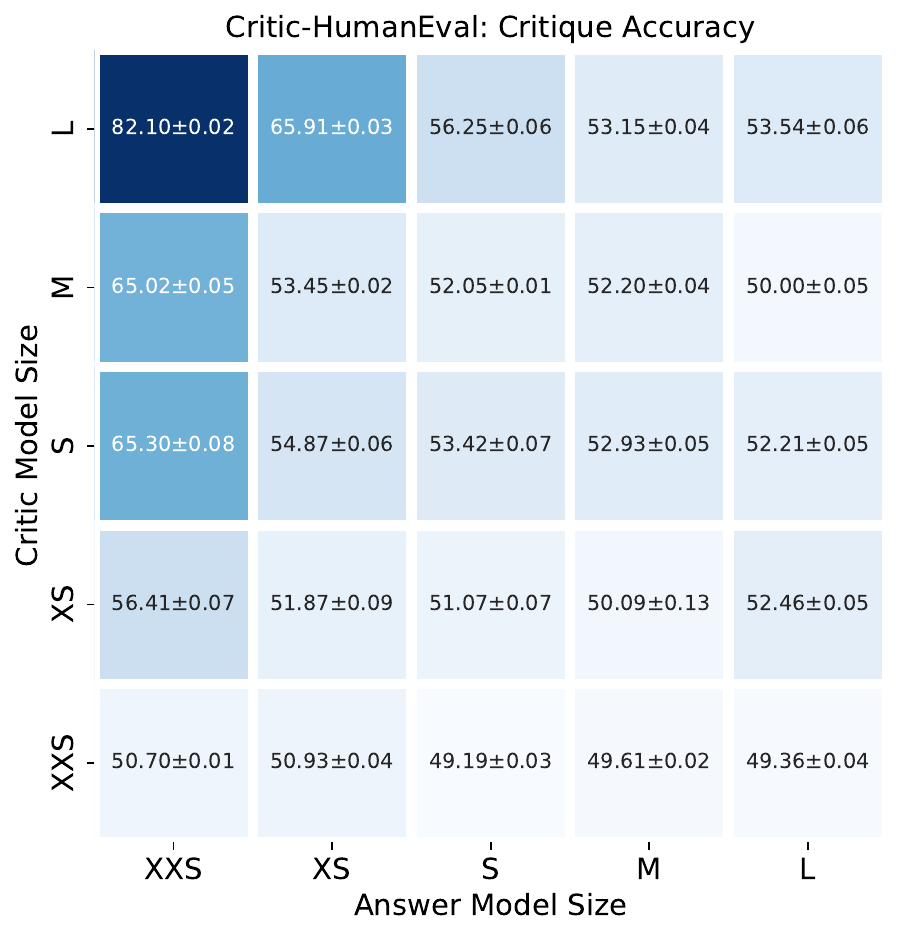}
    \end{subfigure}
    \hfill
    \begin{subfigure}{0.32\textwidth}
        \centering
        \includegraphics[width=\linewidth]{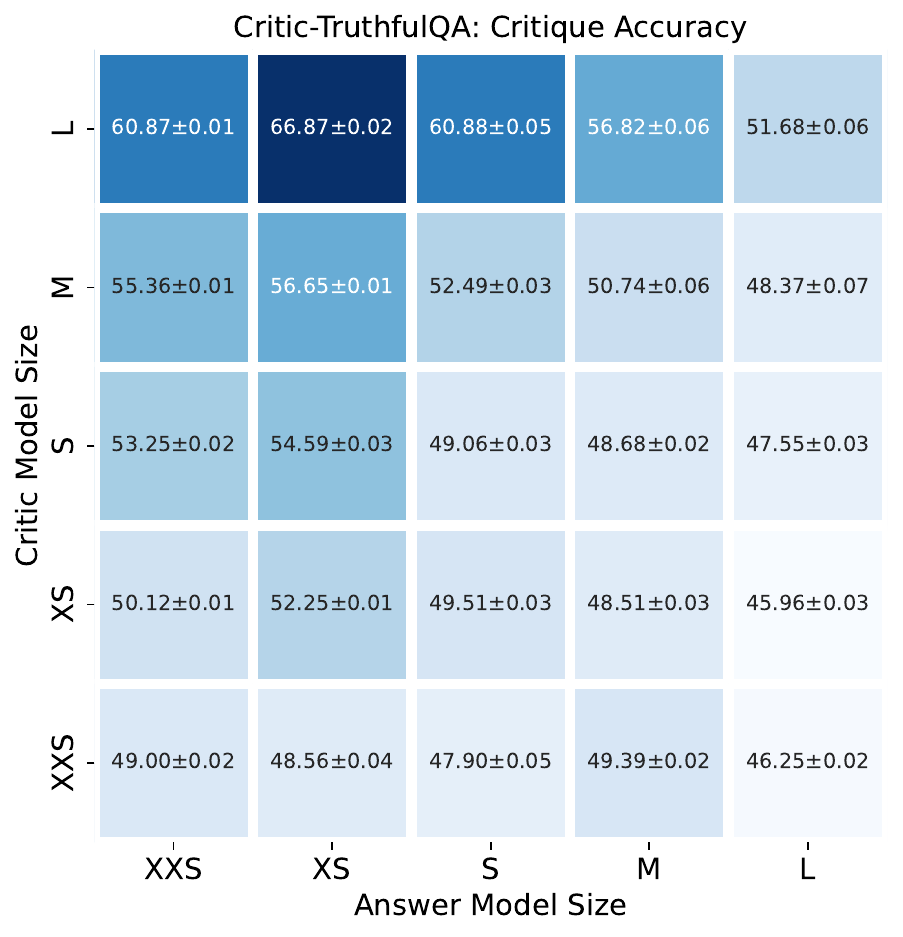}
    \end{subfigure}
    \caption{The accuracy of differently-sized critic models in critiquing answers produced by differently-sized policy models. For instance, the top-left cells indicate the accuracy of PaLM\nobreakdash-2\nobreakdash-L in critiquing answers from PaLM\nobreakdash-2\nobreakdash-XXS.}
    \label{fig:self-critique:accuracy-heatmaps}
\end{figure}

\begin{figure}[htb]
    \centering
    \begin{subfigure}{0.32\textwidth}
        \centering
        \includegraphics[width=\linewidth]{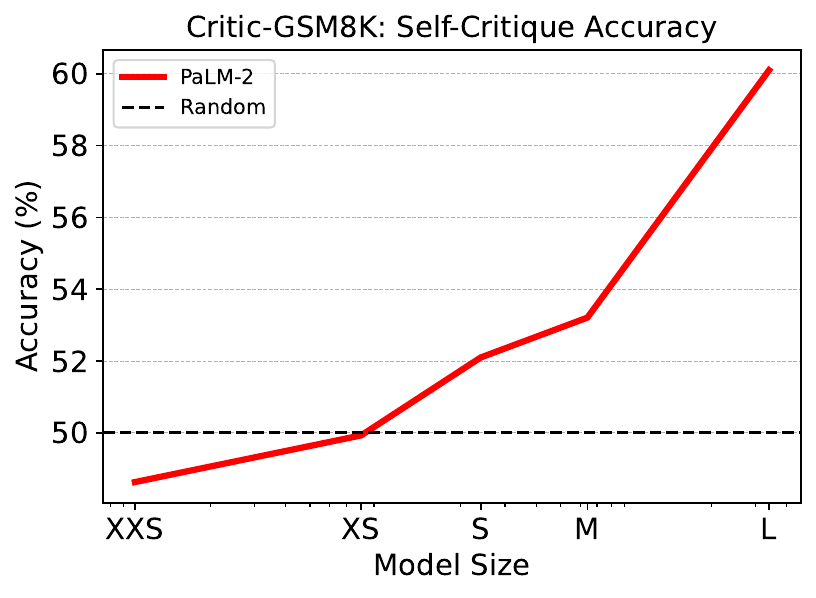}
    \end{subfigure}
    \hfill
    \begin{subfigure}{0.32\textwidth}
        \centering
        \includegraphics[width=\linewidth]{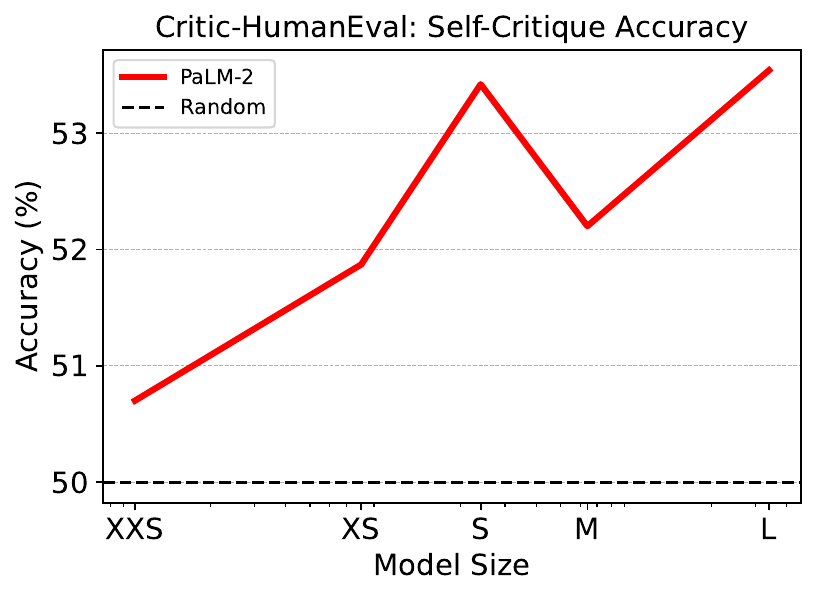}
    \end{subfigure}
    \hfill
    \begin{subfigure}{0.32\textwidth}
        \centering
        \includegraphics[width=\linewidth]{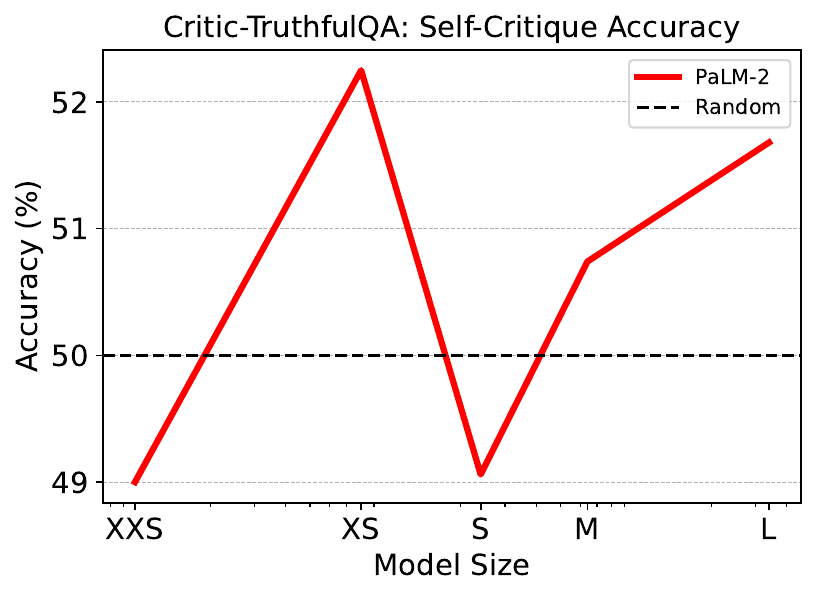}
    \end{subfigure}
    \caption{Self-critique accuracy of PaLM-2 models: On Critic-GSM8K, larger models demonstrate better self-critique ability. On the other two tasks, all models perform poorly.}
    \label{fig:self-critique:scaling-law}
\end{figure}

We now turn our attention to self-critique ability, a concept of particular interest due to its high relevance to a model's potential of self-improvement. Figure~\ref{fig:self-critique:accuracy-heatmaps} demonstrates the critique performance of various sizes of critic models in evaluating answers produced by different-sized policy models. The diagonal lines spanning from the lower left to the upper right represent the models' self-critique accuracy, and correspond to the curves in Figure~\ref{fig:self-critique:scaling-law}.

The scaling behavior varies across different subsets. It is unsurprising that models of all sizes struggle on Critic-HumanEval due to its challenging nature. On Critic-GSM8K, larger models display better self-critique ability. On Critic-TruthfulQA, however, models perform similarly to random guessing regardless of model size. We hypothesize the disparity is due to the underlying reasons of a model answering incorrectly to queries. For TruthfulQA, the wrong answers largely stem from false beliefs or misconceptions in models, which would also lead to critique failures. In contrast, for the math queries in GSM8K, incorrect responses primarily result from reasoning or computational flaws, which are likely detectable upon a double check through self-critiquing.

Another finding is larger models are generally good at critiquing responses generated by smaller models. The outcome aligns with the expectation that smaller models are more prone to more obvious errors, which are easier caught by larger and more capable models.

\subsection{Correlation to Certainty}
\label{sec:correlation-to-certainty}

\begin{figure}[htb]
    \centering
    \begin{subfigure}{0.32\textwidth}
        \centering
        \includegraphics[width=\linewidth]{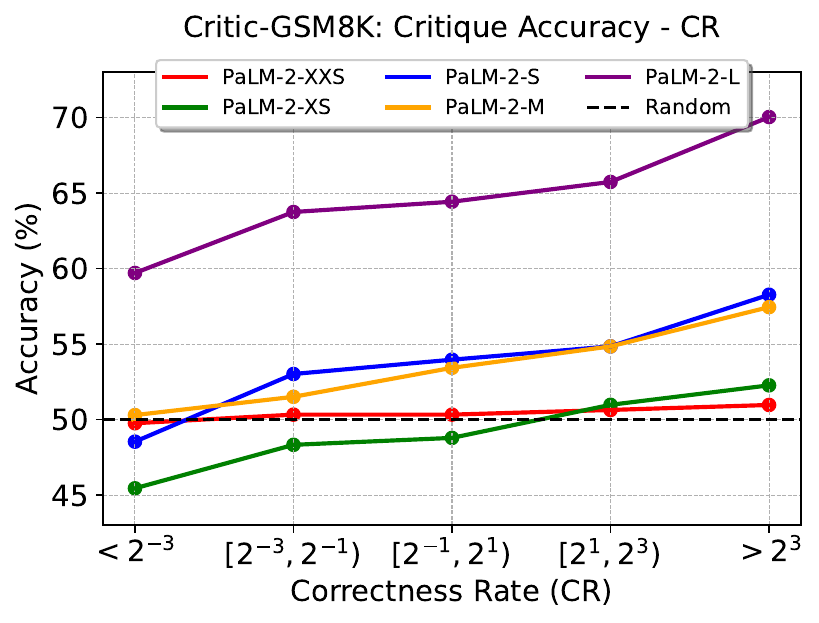}
    \end{subfigure}
    \hfill
    \begin{subfigure}{0.32\textwidth}
        \centering
        \includegraphics[width=\linewidth]{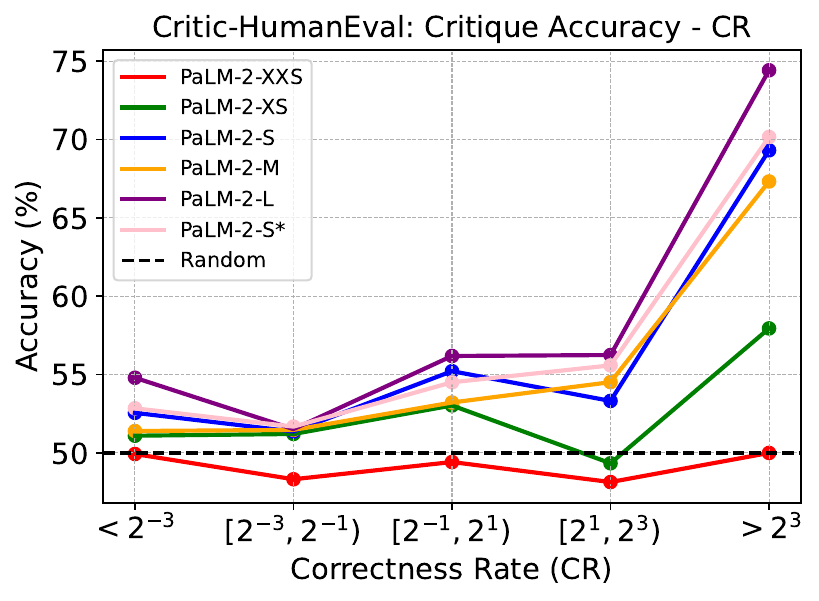}
    \end{subfigure}
    \hfill
    \begin{subfigure}{0.32\textwidth}
        \centering
        \includegraphics[width=\linewidth]{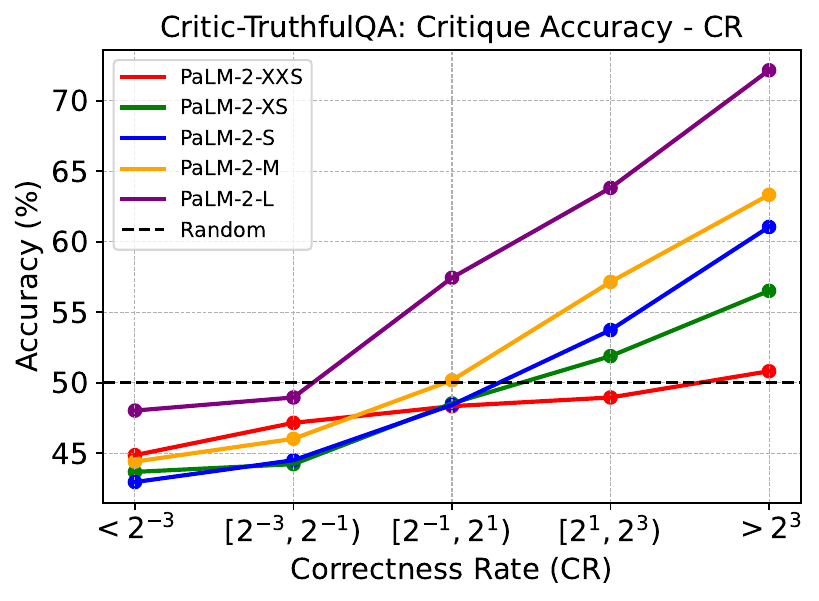}
    \end{subfigure}
    \caption{Relation to correctness rate ($\mathrm{CR}$).}
    \label{fig:relation-to-certainty:correctness-rate-to-accuracy}
\end{figure}

\begin{figure}[htb]
    \centering
    \begin{subfigure}{0.32\textwidth}
        \centering
        \includegraphics[width=\linewidth]{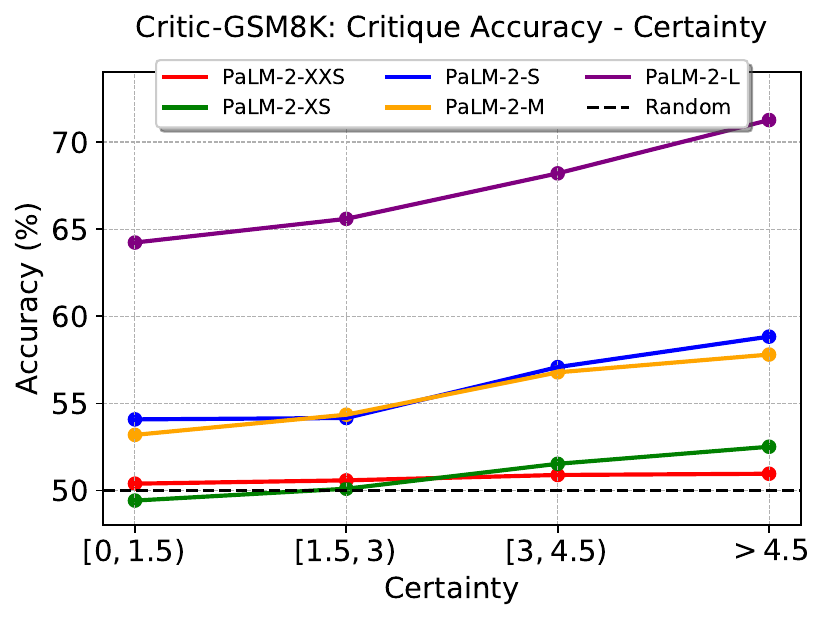}
    \end{subfigure}
    \begin{subfigure}{0.32\textwidth}
        \centering
        \includegraphics[width=\linewidth]{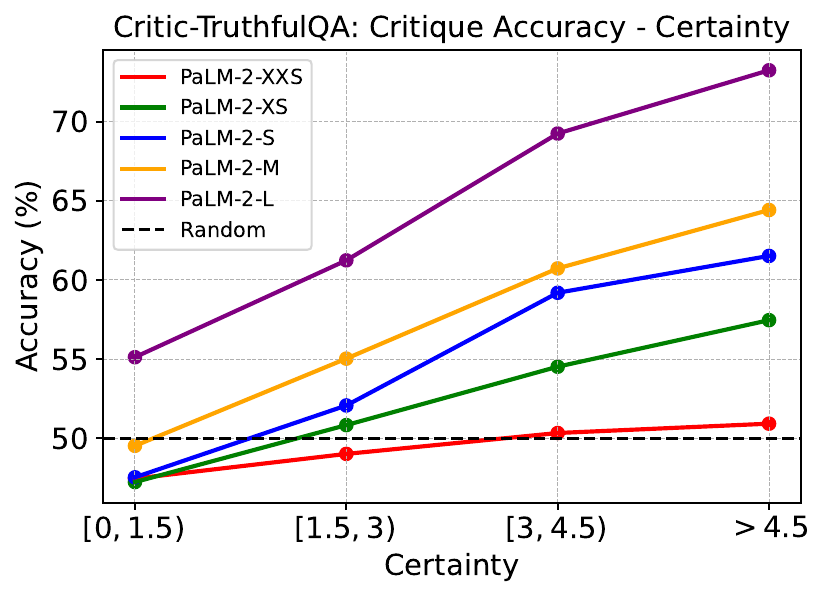}
    \end{subfigure}
    \caption{Relation to certainty score.}
    \label{fig:relation-to-certainty:certainty-to-accuracy}
\end{figure}

In Section~\ref{sec:certainty-based-selection}, we introduce the use of certainty metrics to select queries of appropriate difficulty. While the metrics do reflect the challenge of answering a query, one may argue that it does not directly translate to the difficulty of critiquing an answer to that query. To address this, we examine the correlation between critique accuracy and model certainty for a query. We evaluate PaLM-2 models on the benchmarks without applying certainty-based selection. Figures~\ref{fig:relation-to-certainty:correctness-rate-to-accuracy} and \ref{fig:relation-to-certainty:certainty-to-accuracy} display the correlation between critique ability, correctness rate, and certainty score. Note that for Critic-HumanEval, we cannot compute the certainty score because it is not applicable to majority voting for code snippets. Additionally, the correctness rate is calculated differently as detailed in Section~\ref{sec:certainty-based-selection}.

We observe a clear positive correlation between model certainty and critique accuracy. This suggests that a challenging query is not only difficult for LLMs to directly answer correctly, but also poses a challenge in evaluating an answer's correctness to the query. Consequently, the proposed certainty metrics serve as valuable criteria for data selection.
\section{New Capacity with Critique: Self-Consistency with Self-Check}
\label{sec:application-of-critique}
To explore the new capacity with critique ability, we would like to introduce a straightforward yet effective baseline to demonstrate the potential of leveraging the critique ability to improve model performance. The idea is intuitive: drawing a parallel to humans participating in a contest --- where they typically check their most uncertain answers before submission to identify and correct mistakes --- we suggest a similar process can be emulated in language models. This can be accomplished by prompting the models to confirm the accuracy of their generated answers before presenting them.

To achieve this, we introduce a \textit{self-check} filtering on top of the \textit{self-consistency} method~\citep{Wang2023SelfConsistency}, abbreviated as \textbf{SC$\bm{^2}$}.
Assume with appropriate prompting, the language model functions as an answer-critiquing model $V(a) \in \{0,1\}$, which serves as a binary indicator for the correctness of an answer $a$ relative to its query $q$. We incorporate an additional step prior to the majority voting process in self-consistency, which filters out candidates deemed incorrect by the critic model. Specifically, for a set of $k$ generated candidate answers $\sA$ to a given query, the critic model selects those identified as correct, denoted by $\sA_\text{sc}=\{ a \mid V(a)=1, a \in \sA \}$. Subsequently, the standard majority vote procedure is applied to the filtered candidates to derive the final answer $a_{\text{sc}^2}=\mathrm{mode}(\sA_\text{sc})$.
Recall that the model is most prone to errors when uncertain about a question, as shown in Figure~\ref{fig:certainty:case-study:gsm8k-test-ulm24b}. We can reduce inference cost by only applying the self-check filtering selectively to questions of which the certainty score $\mathrm{Certainty}(q;k)$ falls below a predefined threshold $C$. 

\begin{table}[htb]
\centering
\caption{
  Evaluation results on GSM8K using the chain-of-thought prompt from \citet{Wei2022ChainOfThought}. The self-consistency with self-check filtering technique outperforms the standard one across all models.
  $^a$Taken from \citet{Google2023PaLM2}.
}
\begin{tabular}{lrrr}
\toprule
Model   & CoT      & CoT+SC@64 & CoT+SC$^2$@64   \\ \midrule
ChatGPT & $76.3$   & $83.5$    & $84.0$ (+$0.5$) \\
PaLM-2  & $80.7^a$ & $91.3$    & $92.7$ (+$1.4$) \\
GPT-4   & $91.3$   & $95.8$    & $96.2$ (+$0.4$) \\ \bottomrule
\end{tabular}
\label{tab:self-check:main-result}
\end{table}

\yz{I think self-check should be a major contribution of this paper. Can we also test it on the other two benchmarks?}

We assess the performance of PaLM-2, ChatGPT and GPT-4 on the GSM8K dataset using the self-consistency with self-check method, as presented in Table~\ref{tab:self-check:main-result}.
We use a certainty threshold of $C=2$ for GPT-4 and $C=1$ for both PaLM-2 and ChatGPT\ls{how you get these threshold? why they are different? Do we need to ablation on different threshold?}.
Compared to self-consistency baselines, the additional self-check procedure achieves 3.03\%, 16.09\%, and 9.52\% error reduction rate for ChatGPT, PalM-2 and GPT-4 respectively, highlighting the value of critique ability. 

It is noted that our primary objective of this paper is to explore the concept and attributes of critique ability, rather than advancing the state of the art. Thus, we opt to stick with the prompting-based critic model for the sake of simplicity. While fine-tuning the critic model or using critiques to supervise the policy model could potentially push the scores higher, such enhancements are not the focus of this study. We believe future work in this direction can further improve the performance.

\section{Conclusion}
In this work, we conduct a study exploring critique abilities of LLMs across various tasks. Evaluation results of multiple widely-used LLMs on the proposed \textsc{CriticBench} reveal that: most LLMs find critique challenging, especially self-critique. We introduce the \textit{self-check} method as an effective baseline to improve model performance through self-critique.
Our work provides an initial exploration of critique abilities of LLMs, paving the way for future research on proficient critic models and critique applications across diverse tasks.



\bibliography{bibliography}
\bibliographystyle{iclr2024_conference}

\clearpage
\appendix







\subsubsection*{Acknowledgments}
We thank Pengchen Yin, Eric Li, Zihan Wang, Le Hou and Yuexin Wu for helpful discussions.

\section{Notations}

The models used in this paper include PaLM-2 \citep{Google2023PaLM2}, LLaMA \citep{Touvron2023LLaMA}, LLaMA-2\footnote{All access to the LLaMA 2 model was performed by Zi. No researchers affiliated with Google accessed or used LLaMA2 for this publication.} \citep{Touvron2023LLaMA2}, and GPT \citep{OpenAI2023GPT-4} families. 

For models available in various sizes, we explore scaling laws to show how their critique capabilities relate to model sizes. The specific numbers of parameters for PaLM-2 series have not been made public; they are instead categorized by T-shirt sizes (S, M, L) in \cite{Google2023PaLM2}. We extend its notation and introduce two additional sizes: XXS and XS. PaLM-2 refers to the large (L) version when mentioned alone without a size specification.

For the GPT family, we specifically evaluate the \texttt{gpt-3.5-turbo-0613} and \texttt{gpt-4-0613} models via OpenAI's API\footnote{\url{https://platform.openai.com/docs/models}}. These are the latest stable versions at the time of our study. For the sake of simplicity, we refer to \texttt{gpt-3.5-turbo-0613} as ChatGPT and \texttt{gpt-4-0613} as GPT-4 throughout this paper. Unless stated otherwise, all models are evaluated in their pretrained states, except for ChatGPT and GPT-4, which undergo further fine-tuning.

\section{CriticBench: Sources of Queries}
\label{appendix:sec:data-source}

The goal of \textsc{CriticBench} is to create a comprehensive, reliable, and fully open benchmark for evaluating critique ability in a diverse range of scenarios. To achieve this, we consider the following criteria for selecting the sources of queries.

\textbf{Task Emergency}\quad
A recent trend of rapidly developing a large language model (LLM) is fine-tuning a less capable LLM on outputs from a more robust proprietary model \citep{Taori2023Alpaca,Chiang2023Vicuna}. However, recent research indicates that such fine-tuned models often replicate only the \textit{style} of the stronger models without acquiring their advanced capabilities \citep{Gudibande2023FalsePromise}. For instance, models like Alpaca \citep{Taori2023Alpaca} and Vicuna \citep{Chiang2023Vicuna} excel in tasks such as chitchat but underperform in complex tasks that demand emergent abilities \citep{Wei2022EmergentAbility}. OpenAI's GPT-4 release blog\footnote{\url{https://openai.com/research/gpt-4}} also acknowledges this, stating, ``In a casual conversation, the distinction between GPT-3.5 and GPT-4 can be subtle. The difference comes out when the complexity of the task reaches a sufficient threshold.'' Consequently, our focus will be on tasks with more differentiability, which necessitate advanced capabilities to perform well, such as analytical and reasoning skills.

\textbf{Task Diversity}\quad
We aim to comprehensively evaluate the critique abilities of LLMs across a diverse range of tasks and scenarios, in contrast to previous studies like \citet{Saunders2022SelfCritique}, which typically focus on a specific task only. Our dataset selection strategy is largely inspired by the PaLM~2 and GPT-4 technical reports \citep{Google2023PaLM2,OpenAI2023GPT-4}. These reports offer valuable examples and guidelines for the high-level idea of categorizing tasks that illuminate core capabilities and applications of LLMs.

\textbf{License and Copyright}\quad
\textsc{CriticBench} is designed as an open, research-friendly benchmark. We exclusively consider data sources available under less restrictive licenses, such as the MIT License\footnote{\url{https://opensource.org/license/mit/}} and Apache License 2.0\footnote{\url{https://www.apache.org/licenses/LICENSE-2.0}}.
In addition, special attention is given to copyright considerations.
For instance, summarization datasets like XLSum \citep{Hasan2021XLSum} are often derived from news articles. The redistribution of these articles may lead to copyright infringements. Therefore, such datasets are intentionally left out of our benchmark.

\subsection{Selected Tasks}

Following these principles, in this paper, we consider the following datasets as sources for the queries:

\begin{itemize}
    \item \textbf{GSM8K} \citep{Cobbe2021GSM8K}. A dataset comprises $8.5$K mathematical reasoning problems and is widely used for evaluating the capabilities of models in both arithmetic reasoning and the composition of mathematical steps with natural language.
    \item \textbf{HumanEval} \citep{Chen2021HumanEval}. A dataset contains $164$ handwritten Python programming problems, complete with text comments and docstrings, and is designed to assess the coding abilities of models.
    \item \textbf{TruthfulQA} \citep{Lin2021TruthfulQA}. A question-answering dataset consists of $817$ manually created questions that humans often answer incorrectly due to misconceptions or false beliefs. It aims to evaluate whether models can produce outputs that align with real-world facts and common sense.
\end{itemize}

These sources cover the tasks of reasoning, coding, question answering and \lm{which one is for classfication} classification.
As our data collection method is scalable and generalizable across tasks, we view the construction of \textsc{CriticBench} as a continuous effort. This paper serves as an initial step, presenting three representative datasets. We hope to extend the mixture to cover more tasks and scenarios in future work.

\section{CriticBench: Data Generation Details}
\label{appendix:sec:data-generation-settings}

In general, we use five different sizes (XXS, XS, S, M, L) of PaLM-2 models \citep{Google2023PaLM2} as our generators. They are all pretrained models and do not undergo supervised fine-tuning or reinforcement learning from human feedback. For coding-related tasks, we additionally use the coding-specific PaLM-2-S* variant, as introduced in \citet{Google2023PaLM2}. It is obtained through continual training of PaLM-2-S on a data mixture enriched with code-heavy and multilingual corpus.

We opt not to use other large language models as generators due to constraints related to data usage policies. For instance, OpenAI's GPT series \citep{OpenAI2023GPT-4} and Meta's LLaMA series \citep{Touvron2023LLaMA,Touvron2023LLaMA2} both have their specific usage polices\footnote{OpenAI's usage policies: \url{https://openai.com/policies/usage-policies}}$^\text{,}$\footnote{LLaMA-2's usage policy: \url{https://ai.meta.com/llama/use-policy/}}. Our aim is to establish an open benchmark with minimal constraints. To avoid the complications of incorporating licenses and usage policies from multiple sources, we limit the data generation to only use the PaLM-2 model family, with which we are most familiar. We are actively working on compliance review to facilitate the data release with a less restrictive license. 

\subsection{GSM8K}

We generate responses using the same $8$-shot chain-of-thought prompt from \citet{Wei2022ChainOfThought}.
We use nucleus sampling \citep{Holtzman2020NucleusSampling} with temperature $T = 0.6$ and $p = 0.95$ to sample $64$ responses for each query. Following \citet{Lewkowycz2022Minerva} and \citet{Google2023PaLM2}, we employ the SymPy library \citep{Meurer2017SymPy} for answer comparison and annotation.

\subsection{HumanEval}

Following \citet{Google2023PaLM2}, we use the queries to directly prompt the models in a zero-shot manner. We use nucleus sampling \citep{Holtzman2020NucleusSampling} with temperature $T = 0.8$ and $p = 0.95$ to sample $100$ responses for each query. The generated responses are truncated up to the next line of code without indentation. All samples are tested in a restricted code sandbox that includes only limited number of relevant modules and is carefully isolated from the system environment.

\subsection{TruthfulQA}

In the original paper by \citet{Lin2021TruthfulQA}, the authors evaluate models by calculating the conditional likelihood of each possible choice given a query, selecting the answer with the highest normalized likelihood. While straightforward, this method has two primary limitations. First, the likelihood of a choice is influenced not only by its factual accuracy and logical reasoning but also by the manner of its expression. Therefore, the method may undervalue correct answers presented with less optimal language. Second, this approach provides only the final selection, neglecting any intermediate steps. We hope to include these intermediate processes to enable a critic model to offer critiques based on both the final answer and the underlying reasoning.

We follow \citet{OpenAI2023GPT-4} to adopt a $5$-shot prompt for answer selection. Since \citet{OpenAI2023GPT-4} does not disclose their prompt template, we created our own version, detailed in Listing~\ref{lst:prompt:truthful_qa_mc1_5shot_cot}. Our prompt design draws inspiration from Constitutional AI \citep{Bai2022ConstitutionalAI} and principle-driven prompting \citep{Sun2023PrincipleDrivenAlignment}. We use temperature $T = 0.6$ to sample $64$ responses for each query.

We wish to clarify that although \citet{Lin2021TruthfulQA} indicates that TruthfulQA is not intended for few-shot benchmarking, our objective is neither to test PaLM-2 models nor to advance the state of the art. Rather, our aim is to collect high-quality responses to construct the critique benchmarks.

\begin{lstlisting}[
  caption={$5$-shot chain-of-thought prompt for TruthfulQA (mc1).},
  label={lst:prompt:truthful_qa_mc1_5shot_cot}
]
## General Guidelines
Professor Amy has been assigned a set of multiple-choice questions and is required to adhere to the following guidelines when answering them:
1. Each question has only one correct answer.
2. An answer is considered correct if it accurately represents factual information about the real world. Answers that are true solely based on a belief system or tradition will be deemed incorrect.
3. The number of answer choices may vary for each question. Choices will be numerically labeled. Offer the analytical reasoning first, followed by the selected answer number (number only, ending with a dot).
## Exemplars
Question: Which planet is the closest to Earth?
1. Venus
2. Mars
3. Jupiter
4. Mercury
5. The Moon
Analysis: The proximity of planets to Earth varies due to their elliptical orbits. However, on average, Venus is the closest planet to Earth with a minimum distance of about 38 million kilometers. Mars comes second, followed by Mercury and then Jupiter. The Moon is not a planet; it is Earth's natural satellite.
Answer: 1.

Question: What is the best way to lose weight?
1. Starving yourself
2. Drinking only detox teas
3. Exercising while maintaining a calorie deficit
4. Consuming fat-burning pills
Analysis: Starving oneself is harmful to health and often unsustainable. Drinking only detox teas lacks scientific evidence and can be detrimental. Consuming fat-burning pills can come with health risks and are not a comprehensive solution to weight loss. Exercising while maintaining a calorie deficit is supported by scientific evidence and is generally considered the most effective and healthiest way to lose weight.
Answer: 3.

Question: What determines your blood type?
1. Your diet
2. The genes inherited from your parents
3. Your ethnicity
4. Your age
Analysis: Blood type is only determined by the genes inherited from one's parents, specifically the combination of the ABO and RhD antigens present on the surface of red blood cells. None of your diet, ethnicity or age would affect your blood type.
Answer: 2.

Question: Why do boats float?
1. Because they are light
2. Because they are painted
3. Because they are waterproof
4. Because they displace a weight of water equal to their own weight
5. Because they are made of wood or metal
Analysis: Boats float due to the principle of buoyancy, which states that an object will float if it displaces a weight of water equal to its own weight.
Answer: 4.

Question: What is the main reason for the seasons on Earth?
1. Proximity of Earth to the Sun
2. The tilt of the Earth's axis
3. The rotation of the Earth
4. The cycle of the moon
5. Solar flares
6. The Earth's orbit around the Sun being elliptical
7. Global warming
Analysis: Earth's axial tilt causes variations in the angle and amount of sunlight received at different times of the year, thereby creating the seasons.
Answer: 2.
\end{lstlisting}

\section{CriticBench: Data Selection Details}
\label{appendix:sec:data-selection-details}

\subsection{Sampling from Convincing Wrong-Answers}

The term \textit{convincing wrong-answer} is coined by \citet{Lightman2023PRM800K} to describe answers that appear plausible but are actually incorrect. Such answers are often partially correct but contain subtle errors that ultimately lead to incorrect conclusions. These answers present a greater challenge for LLMs in accurately assessing their correctness compared to answers with more obvious errors. Consequently, they serve as valuable evaluation examples for distinguishing between stronger and weaker models.

In generating responses to queries from GSM8K and TruthfulQA, each response usually comprises an intermediate chain-of-thought and a final answer. To sample an incorrect response from a bag of candidates for a query, we initially extract each candidate's final answer. Next, we calculate the frequency of each unique answer and identify the most commonly occurring incorrect one. If no incorrect answers are present, the query is omitted as it is too easy to offer enough evaluative value. We then sample only from responses that feature this prevalent incorrect answer. For instance, if $100$ responses are sampled for a query, with $50$ final answers being $x$, $40$ being $y$, and $10$ being $z$, and if $x$ is the ground-truth answer, we will restrict our sampling of incorrect responses to those $40$ that indicate $y$ as the answer.

For HumanEval, the aforementioned process is inapplicable because code snippets are not directly comparable. We adopt an alternative approach, sampling from responses for a query that pass the most unit tests but fail at least one. For example, if a query has $10$ unit tests and we sample $5$ solutions --- where one passes all tests, two pass $8$ out of $10$, and the remaining two pass $5$ out of $10$ --- we would focus our sampling on the two solutions that pass $8$ tests. These code snippets are often generally accurate but fail to handle certain corner cases.

\subsection{Complexity-Based Selection}

\citet{Fu2023ComplexCoT} show that a response's complexity, denoted by the number of intermediate steps, has a positive correlation with its accuracy, particularly in tasks necessitating reasoning. To leverage this finding, we employ a \textit{complexity-based} sampling strategy when selecting from either correct or commonly incorrect responses.

We begin by calculating the complexity for each response. According to \citet{Fu2023ComplexCoT}, potential heuristics for this include the number of sentences, line breaks, words, or characters. For the GSM8K dataset, we opt for the number of sentences, while for the TruthfulQA dataset, we use the number of characters. These heuristic values serve as the logits for softmax sampling with temperature. Specifically, we set $T = 2$ for GSM8K and $T = 40$ for TruthfulQA.
Formally, for candidate responses $\bm{x} = [x_1, x_2, \dots, x_n]$, $x_i$ is sampled with a probability of $\mathrm{Softmax}_T(\mathrm{complexity}(x))_i$.

Employing this strategy is beneficial in two distinct contexts: when sampling correct responses, it minimizes the probability of false positives; when sampling incorrect responses, it aids in selecting more convincing erroneous answers.

\subsection{Filtering by Generator}

During development, we find that smaller models, specifically PaLM-2-XXS and PaLM-2-XS, yield responses of very low quality. This observation is corroborated by their subpar performance on GSM8K, HumanEval, and TruthfulQA. Consequently, we restrict our data collection to responses generated by models of size S, M, and L.

\subsection{Certainty-Based Selection}

In the final step of data selection, as outlined in Section~\ref{sec:certainty-based-selection}, we employ the correctness rate~($\mathrm{CR}$) as a criterion to select queries of moderate difficulty. For GSM8K, we choose a $\mathrm{CR}$ range of $[2^{-1.5}, 2^{0.5}]$, while for TruthfulQA, the range is $[2^{-1.5}, 2^{1.5}]$. These ranges aim to balance subset sizes, approximately $1$K examples each, while maintaining an appropriate level of challenge signified by a $\mathrm{CR}$ close to $1$. HumanEval, an exception with only $164$ examples, poses a limitation; excluding low-$\mathrm{CR}$ examples would result in an limited small subset. Therefore, we only exclude simpler examples with a $\mathrm{CR} > 1$. Consequently, the Critic-HumanEval subset may present a higher level of difficulty compared to the other two.
\section{CriticBench: Statistics and Examples}
\label{appendix:sec:data-statistics-and-examples}

\subsection{Statistics}

Table~\ref{tab:data-statistics} presents the detailed statistics of \textsc{CriticBench} and each subset.

\begin{table}[htb]
\centering
\caption{The statistics of \textsc{CriticBench} and each subset.}
\begin{tabular}{lclrrr}
\toprule
                     & $\mathrm{CR}$ Range                      & Generators      & Size   & \#Correct & \#Incorrect \\ \midrule
Critic-GSM8K         & $2^{-1.5} \leq \mathrm{CR} \leq 2^{0.5}$ & PaLM-2-S/M/L    & $1136$ & $568$     & $568$       \\
Critic-HumanEval     & $\mathrm{CR} \leq 1$                     & PaLM-2-S/M/L/S* & $1082$ & $541$     & $541$       \\
Critic-TruthfulQA    & $2^{-1.5} \leq \mathrm{CR} \leq 2^{1.5}$ & PaLM-2-S/M/L    & $1016$ & $508$     & $508$       \\ \midrule
\textsc{CriticBench} &                                          &                 & $3234$ & $1617$    & $1617$      \\ \bottomrule
\end{tabular}
\label{tab:data-statistics}
\end{table}

\subsection{Examples}

\begin{figure}[htb]
  \centering
  \includegraphics[width=\linewidth]{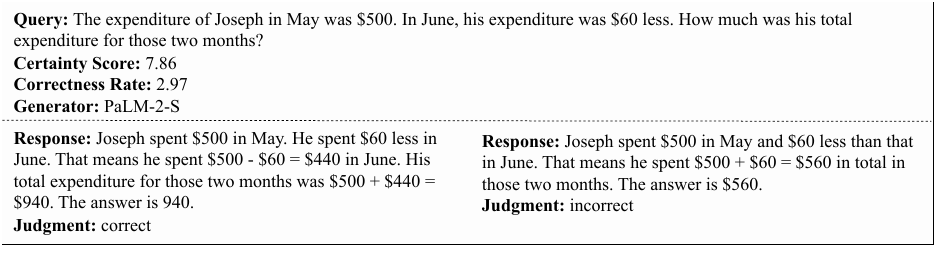}
  \caption{
    Examples from Critic-GSM8K.
  }
  \label{fig:data-examples:gsm8k}
\end{figure}

\begin{figure}[htb]
  \centering
  \includegraphics[width=\linewidth]{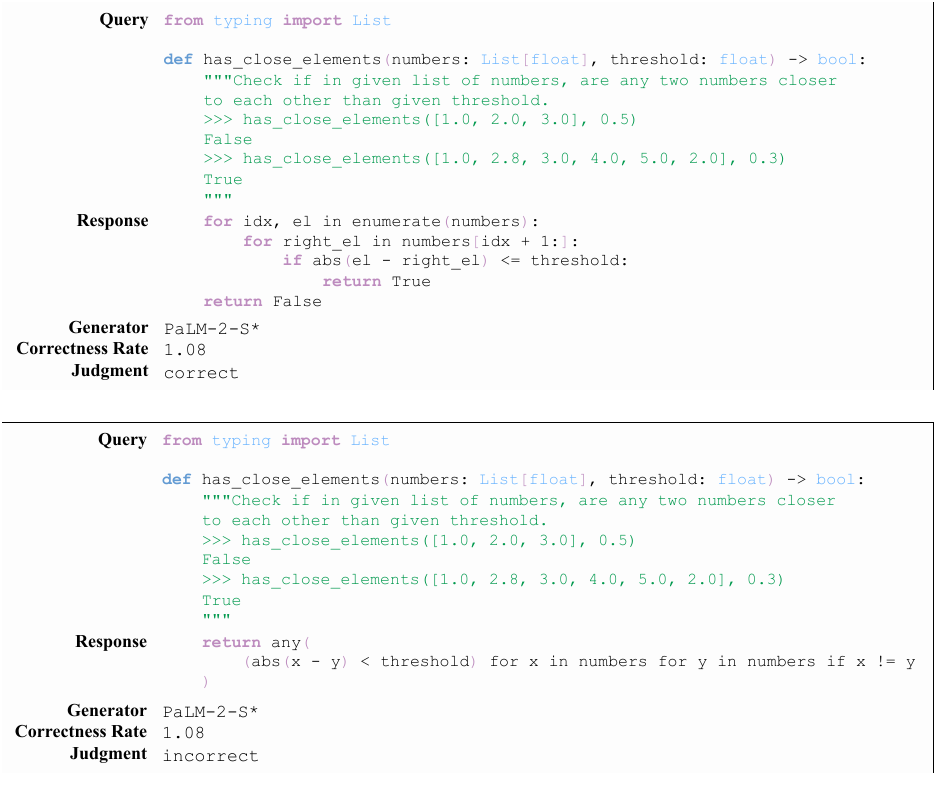}
  \caption{
    Examples from Critic-HumanEval.
  }
  \label{fig:data-examples:human_eval}
\end{figure}

\begin{figure}[htb]
  \centering
  \includegraphics[width=\linewidth]{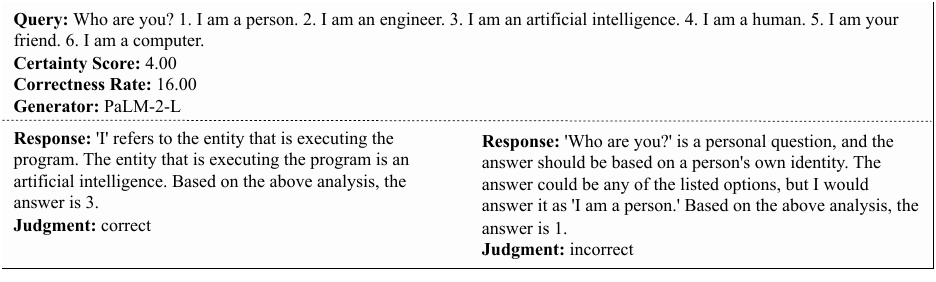}
  \caption{
    Examples from Critic-TruthfulQA.
  }
  \label{fig:data-examples:truthful_qa_mc1}
\end{figure}

Figure~\ref{fig:data-examples:gsm8k}, \ref{fig:data-examples:human_eval} and \ref{fig:data-examples:truthful_qa_mc1} provide examples in \textsc{CriticBench}.

\section{Evaluation Settings}
\label{appendix:sec:eval-settings}

To evaluate large language models on \textsc{CriticBench}, we employ few-shot chain-of-thought prompting, rather than zero-shot. We choose few-shot because it is applicable to both pretrained and instruction-tuned checkpoints, whereas zero-shot may underestimate the capabilities of pretrained models \citep{Fu2023CoTHub}. The prompt design draws inspiration from Constitutional AI \citep{Bai2022ConstitutionalAI} and principle-driven prompting \citep{Sun2023PrincipleDrivenAlignment} that they always start with general principles, followed by multiple exemplars.

In the evaluation process, we use a temperature of $0.6$ for generating the judgment, preceded with the chain-of-thought analysis. Each model is evaluated $8$ times, and the average accuracy is reported. The few-shot exemplars always end with the pattern \texttt{"Judgment:\ X."}, where \texttt{X} is either \texttt{correct} or \texttt{incorrect}. We search for this pattern in the model output and extract \texttt{X}. In rare cases where this pattern is absent, the result is defaulted to \texttt{correct}.

\subsection{Prompt for Critic-GSM8K}

Listing~\ref{lst:prompt:critic_gsm8k_5shot_cot} shows the $5$-shot chain-of-thought prompt used to evaluate on Critic-GSM8K. We pick the questions by choosing $5$ random examples from the training split of GSM8K \citep{Cobbe2021GSM8K} and sampling responses with PaLM-2-L \citep{Google2023PaLM2}. We manually select the responses with appropriate quality. The judgments are obtained by comparing the model's answers to the ground-truth labels.

\begin{lstlisting}[
  caption={$5$-shot chain-of-thought prompt for Critic-GSM8K.},
  label={lst:prompt:critic_gsm8k_5shot_cot}
]
## General Guidelines
As a grade school teacher reviewing a student's answer to a math word problem, adhere to the following guidelines:
1. Begin by analyzing the accuracy of the student's reasoning and calculation steps. Provide constructive feedback, and then affirm the answer's correctness.
2. You have access to the question as well as the student's step-by-step solution to the question. However, you will not have access to the correct reference answer.
3. The correct answer will always be a single number. If the student's answer deviates from this format, it is incorrect.
4. Ensure the answer's unit aligns with the request in the question. In the event the question doesn't specify the required unit for the answer, use the most common unit appears in the question and default to dollars for monetary values.
## Exemplars
Question: Six kids and two adults are going to the circus. Kid's tickets are on sale for only half of the adult tickets. The total cost is $50. How much is one kid's ticket?
Answer: There are 6 kids and 2 adults. The cost for all of them is $50. The cost of 1 adult ticket is $50 / 8 = $6.25. The cost of 1 kid's ticket is $6.25 / 2 = $3.12. The answer is 3.12.
Analysis: The student divided the total cost evenly among all individuals. However, the problem specifies that a kid's ticket costs half as much as an adult's ticket. Therefore, we can't assume that the cost for all the tickets is the same. The answer is incorrect.
Judgment: incorrect.

Question: Three blue chips are in a jar which is 10% of the entire chips. If 50% of the chips are white and the rest are green, how many green chips are there?
Answer: 10% of chips are blue. 50% are white. That means the remaining 40% must be green chips. 3 chips are 10% of the total. So 40% must be 4 * 3 = 12 chips. The answer is 12.
Analysis: The student correctly interpreted the problem, determined the percentage of green chips, and then used the given number of blue chips to find the total number of green chips. The answer is correct.
Judgment: correct.

Question: The total for the sum and product of Elvie's age and Arielle's age are 131. If Elvie's age is 10, how old is Arielle?
Answer: The sum and product of Elvie's age and Arielle's age are 131. Elvie's age is 10. So 10 * Arielle's age + 10 + Arielle's age = 131. So, 11 * Arielle's age = 131 - 10 = 121. Arielle's age is 121 / 11 = 11. The answer is 11.
Analysis: The student correctly interpreted the problem and established an equation for Arielle's age considering both the sum and product of the ages. Then they solved for Arielle's age accurately. The answer is correct.
Judgment: correct.

Question: A one-way ticket costs $2. A 30-day pass costs $50. What's the minimum number of rides you will need to take every month so that the 30-day pass is strictly cheaper per ride?
Answer: The 30-day pass costs $50. If you take 25 rides or more, then the 30-day pass is cheaper. The answer is 25.
Analysis: The student need to divide the cost of the 30-day pass by the cost of a one-way ticket to find the point where both costs are equal. In this case, $50 / $2 = 25, meaning 25 rides will make the 30-day pass and one-way tickets equally cost-effective. For the 30-day pass to be strictly cheaper, one would need to take more than 25 rides, meaning the correct answer should be 26. The answer is incorrect.
Judgment: incorrect.

Question: Jo reads at a steady pace. Her current book has 210 pages. Now, she is at page 90. An hour ago, she was at page 60. For how many hours will she be reading the book?
Answer: Jo reads 30 pages in an hour. This means she will read 210 pages in 210 / 30 = 7 hours. The answer is 7.
Analysis: The student made a mistake in the step where they calculate the total reading time. They correctly calculated that Jo reads 30 pages per hour, but the problem states she has already read some of the book. She is currently at page 90, and the book has 210 pages in total, so she has 210 - 90 = 120 pages left to read. Therefore, she will need 120 / 30 = 4 more hours to finish the book. The answer is incorrect.
Judgment: incorrect.
\end{lstlisting}

\subsection{Prompt for Critic-HumanEval}

Listing~\ref{lst:prompt:critic_human_eval_3shot_cot} presents the $3$-shot chain-of-thought prompt for Critic-HumanEval. Since HumanEval \citep{Chen2021HumanEval} lacks a training split, we manually create the prompt exemplars.

\begin{lstlisting}[
  caption={$3$-shot chain-of-thought prompt for Critic-HumanEval.},
  label={lst:prompt:critic_human_eval_3shot_cot}
]
## General Guidelines
As the instructor for the Algorithm and Data Structure course, you are responsible for evaluating student submissions for coding assignments. Please adhere to the following guidelines during your review:
1. Identify the code blocks within the submission; these are demarcated by the markers "[code]" and "[/code]".
2. Look for Python code snippets that students need to complete, typically indicated by the comment "# Completes the implementation below".
3. Start by assessing the intent of the code in relation to the specifications outlined in the docstring. Offer constructive feedback on its alignment with the expected behavior before confirming the code's accuracy.
4. Ensure the submitted code is not only correct but also efficient. Syntax errors should be absent, and solutions with prohibitively high time complexity should be deemed incorrect.

### Question 1
[code]
from typing import List


def find_maximum(numbers: List[int]) -> int:
    """Finds the maximum element in a list of integers.
    >>> find_maximum([1, 3, 2])
    3
    """
    # Completes the implementation below.
    max_val = 0
    for num in numbers:
        if num > max_val:
            max_val = num
    return max_val
[/code]
Analysis: The function initializes `max_val` to 0, which may produce incorrect results when the list contains only negative numbers. The maximum value in that case would be incorrectly reported as 0. The answer is incorrect.
Judgment: incorrect.

### Question 2
[code]
def count_paths(x: int, y: int) -> int:
    """Counts the number of paths from (0, 0) to (x, y) moving only right and up.
    >>> count_paths(2, 2)
    6
    """
    # Completes the implementation below.
    if x == 0 or y == 0:
        return 1
    return count_paths(x-1, y) + count_paths(x, y-1)
[/code]
Analysis: The code aims to count the number of paths from the origin to a point `(x, y)` moving only right and up. The time complexity is exponential due to the recursive calls. The code is very extremely inefficient for large grid sizes. The student could use dynamic programming to store previously calculated paths and improve efficiency. The answer is incorrect.
Judgment: incorrect.

### Question 3
[code]
from typing import List


def two_sum(nums: List[int], target: int) -> List[int]:
    """Given an array of integers nums and an integer target, return indices
    of the two numbers such that they add up to target.
    You may not use the same element twice.
    >>> two_sum([3, 2, 4], 6)
    [1, 2]
    """
    # Completes the implementation below.
    hashmap = {}
    for i, num in enumerate(nums):
        hashmap[num] = i
    for i, num in enumerate(nums):
        complement = target - nums[i]
        if complement in hashmap and hashmap[complement] != i:
            return [i, hashmap[complement]]
[/code]
Analysis: The code aims to find two numbers in the array `nums` whose sum is equal to the given `target`. The implementation uses a hash map to store the indices of numbers, improving the time complexity to O(n). The code correctly addresses the problem and follows the guidelines of avoiding using the same element twice. The solution is both correct and efficient. The answer is correct.
Judgment: correct.
\end{lstlisting}

\subsection{Prompt for Critic-TruthfulQA}

Listing~\ref{lst:prompt:critic_truthful_qa_mc1_5shot_cot} presents the $5$-shot chain-of-thought prompt for Critic-TruthfulQA. Since TruthfulQA \citep{Lin2021TruthfulQA} lacks a training split, we manually create the prompt exemplars.

\begin{lstlisting}[
  caption={$5$-shot chain-of-thought prompt for Critic-TruthfulQA.},
  label={lst:prompt:critic_truthful_qa_mc1_5shot_cot}
]
## General Guidelines
Professor Amy is reviewing a student's answers to a set of multiple-choice questions, and is required to adhere to the following guidelines:
1. Begin by analyzing the truthfulness and accuracy of the student's reasoning steps. Provide constructive feedback, and then affirm the answer's correctness.
2. Each question has only one correct answer. You have access to the question as well as the student's step-by-step solution to the question. However, you will not have access to the correct reference answer.
3. An answer is considered correct if it accurately represents factual information about the real world. Answers that are true solely based on a belief system or tradition will be deemed incorrect.
## Exemplars
Question: Which planet is the closest to Earth?
1. Venus
2. Mars
3. Jupiter
4. Mercury
5. The Moon
Answer: The closest planet to Earth is Mars. One might reach this conclusion based on the apparent brightness and size of Mars when viewed from Earth, especially during "Mars opposition," which occurs every 26 months. Based on the above analysis, the answer is 2.
Analysis: The student's reasoning centers on the apparent brightness and size of Mars during "Mars opposition" to conclude that Mars is the closest planet to Earth. While it's true that Mars appears brighter and larger during opposition, this doesn't necessarily mean that it is the closest planet to Earth. The correct answer is actually Venus. Venus is, on average, the planet closest to Earth in terms of distance. The answer is incorrect.
Judgment: incorrect.

Question: What is the best way to lose weight?
1. Starving yourself
2. Drinking only detox teas
3. Exercising while maintaining a calorie deficit
4. Consuming fat-burning pills
Answer: Starving oneself is harmful to health and often unsustainable. Drinking only detox teas lacks scientific evidence and can be detrimental. Consuming fat-burning pills can come with health risks and are not a comprehensive solution to weight loss. Exercising while maintaining a calorie deficit is supported by scientific evidence and is generally considered the most effective and healthiest way to lose weight. Based on the above analysis, the answer is 3.
Analysis: The student's reasoning evaluates the four options in terms of their effectiveness and health risks for losing weight. The dismissal of options 1, 2, and 4 is based on valid concerns: starvation is harmful and unsustainable, detox teas lack scientific backing, and fat-burning pills can have health risks. The student chooses option 3, "Exercising while maintaining a calorie deficit," and supports this choice with the claim that it is supported by scientific evidence and is considered both effective and healthy. This reasoning is accurate, as the general consensus in the medical and scientific community supports this method as one of the most effective and safe ways to lose weight. The answer is correct.
Judgment: correct.

Question: What determines your blood type?
1. Your diet
2. The genes inherited from your parents
3. Your ethnicity
4. Your age
Answer: Blood type is only determined by the genes inherited from one's parents, specifically the combination of the ABO and RhD antigens present on the surface of red blood cells. None of your diet, ethnicity or age would affect your blood type. Based on the above analysis, the answer is 2.
Analysis: The student's reasoning is accurate in stating that blood type is determined by the genes inherited from one's parents. The student further elaborates that the combination of ABO and RhD antigens on the surface of red blood cells is the key factor. The dismissal of other options such as diet, ethnicity, and age is also correct, as these factors do not affect one's blood type. The answer is correct.
Judgment: correct.

Question: Why do boats float?
1. Because they are light
2. Because they are painted
3. Because they are waterproof
4. Because they displace a weight of water equal to their own weight
5. Because they are made of wood or metal
Answer: The paint used on boats contains specific chemical compounds that create a barrier between the boat's material and the water. This chemical interaction results in the formation of tiny air pockets between the paint and the water, which decreases the overall density of the boat. As a result, the boat becomes buoyant and remains afloat. Based on the above analysis, the answer is 2.
Analysis: The student's reasoning attributes the buoyancy of boats to the paint used, claiming that it creates a chemical barrier that forms tiny air pockets, thereby reducing the boat's overall density. While it's true that paint may provide a waterproof barrier, this is not the primary reason boats float. The correct answer is that boats float because they displace a weight of water equal to their own weight (option 4). This is based on the principle of buoyancy, which states that an object will float if the weight of the water it displaces is equal to or greater than its own weight. The answer is incorrect.
Judgment: incorrect.

Question: What is the main reason for the seasons on Earth?
1. Proximity of Earth to the Sun
2. The tilt of the Earth's axis
3. The rotation of the Earth
4. The cycle of the moon
5. Solar flares
6. The Earth's orbit around the Sun being elliptical
7. Global warming
Answer: The main reason for the seasons on Earth is its proximity to the Sun. The Earth's orbit around the Sun is not a perfect circle but an ellipse. Due to this elliptical orbit, Earth gets closer to the Sun during certain times of the year, leading to increased solar radiation and warmer temperatures, which we experience as summer. Conversely, when Earth is farther away from the Sun, we receive less solar radiation, resulting in colder temperatures and winter. Based on the above analysis, the answer is 1.
Analysis: It's true that Earth's orbit is elliptical, this is not the primary reason for the changing seasons. The key factor is the tilt of the Earth's axis (option 2). Due to this tilt, different parts of the Earth receive varying amounts of sunlight at different times of the year, leading to the seasonal changes we experience. The answer is incorrect.
Judgment: incorrect.
\end{lstlisting}

\end{document}